\title{Learning to detect chest radiographs containing pulmonary lesions using visual attention networks}

\author{Emanuele Pesce$^{1,\footnote{ Present: Warwick Manufacturing Group, University of Warwick, Coventry, UK }}$  
	\and Samuel Withey$^{2,3}$ \and Petros-Pavlos Ypsilantis$^{1}$ \and Robert Bakewell$^{4}$ \and Vicky Goh$^{2,3}$ \and 
	Giovanni Montana $^{1,{}^*,\footnote{Corresponding author, \texttt{g.montana@warwick.ac.uk}}
		}$}
\date{%
	\small
	$^1$Department of Biomedical Engineering, King's College London, London, UK\\%
	$^2$Department of Radiology, Guy's \& St Thomas' NHS Foundation Trust, London, UK\\%
	$^3$Department of Cancer Imaging, King's College London, London, UK\\%
	$^4$Department of Medicine, Imperial College Healthcare NHS Trust, London, UK\\%
}

\documentclass{article}

\usepackage{lineno,hyperref}
\modulolinenumbers[5]

\usepackage{caption}
\usepackage{amsmath}
\usepackage{amssymb}
\usepackage{csquotes}
\usepackage{tabu}
\usepackage{multirow}
\usepackage{makecell}

\usepackage{xcolor}

\usepackage{float}
\usepackage{subfigure}
\usepackage{xstring}
\usepackage{array}
\usepackage{chngcntr}

\def\expaddtocs#1#2{%
	\expandafter\expandafter\expandafter\def\expandafter\expandafter\expandafter
	#1\expandafter\expandafter\expandafter{\expandafter#1#2}}
\def\swapcol#1#2#3\\{%
	\ifnum#1>#2 \def\colinf{#2}\def\colsup{#1}%
	\else \def\colinf{#1}\def\colsup{#2}\fi
	\noexpandarg \let\swaprow\empty \ifnum0=`}\fi
	\expandafter\StrBefore\expandafter[\number\numexpr\colinf-1]{#3&}&[\beforecol]%
	\ifx\beforecol\empty\else\expaddtocs\swaprow{\beforecol&}\fi
	\StrBehind[\colinf]{&#3&}&[\aftercol]%
	\expandafter\StrBefore\expandafter{\aftercol}&[\colA]%
	\expandafter\StrBehind\expandafter{\aftercol}&[\aftercol]%
	\expandafter\StrBefore\expandafter[\number\numexpr\colsup-\colinf-1\expandafter]\expandafter{\aftercol}&[\intercol]%
	\StrBehind[\colsup]{&#3}&[\aftercol]%
	\expandafter\StrBefore\expandafter{\aftercol&}&[\colB]%
	\expandafter\StrBehind\expandafter{\aftercol}&[\aftercol]%
	\expaddtocs\swaprow{\colB&}%
	\ifx\intercol\empty\else\expaddtocs\swaprow{\intercol&}\fi
	\expaddtocs\swaprow{\colA}%
	\ifx\aftercol\empty\else\expaddtocs\swaprow{\noexpand&}\fi
	\expaddtocs\swaprow{\aftercol\\}%
	\ifnum0=`{\fi\iffalse}\fi \swaprow}

\usepackage[round,sort]{natbib}
\usepackage{graphicx}

\begin{document}

\maketitle

		\begin{abstract}
			Machine learning approaches hold great potential for the automated detection of lung nodules on chest radiographs, but training algorithms requires very large amounts of manually annotated radiographs, which are difficult to obtain. 
			{The increasing availability of PACS (Picture Archiving and Communication System), is laying the technological foundations needed to make available large volumes of clinical data and images from hospital archives.}
			Binary labels indicating whether a radiograph contains a pulmonary lesion can be extracted at scale, using natural language processing algorithms. In this study, we propose two novel neural networks for the detection of chest radiographs containing pulmonary lesions. Both architectures make use of a large number of weakly-labelled images combined with a smaller number of manually annotated x-rays. The annotated lesions are used during training to deliver a type of visual attention feedback informing the networks about their lesion localisation performance. The first architecture extracts saliency maps from high-level convolutional layers and compares the inferred position of a lesion against the true position when this information is available; a localisation error is then back-propagated along with the softmax classification error. The second approach consists of a recurrent attention model that learns to observe a short sequence of smaller image portions through reinforcement learning; the reward function penalises the exploration of areas, within an image, that are unlikely to contain nodules. Using a repository of over {430,000} historical chest radiographs, we present and discuss the proposed methods over related architectures that use either weakly-labelled or annotated images only. 
		\end{abstract}

	\section{Introduction}

	{L}ung cancer is the most common cancer worldwide and the second most common cancer in Europe and the USA \citep{Ferlay20131374,cancerUS}. Due to delays in diagnosis, it is typically discovered at an advanced stage with a very low survival rate \citep{cancerUK}. The chest radiograph is the most commonly performed radiological investigation in the initial assessment of suspected lung cancer because it is inexpensive and delivers a low radiation dose. On a chest radiograph, a nodule is defined as a rounded opacity $ \leq $ $3$cm, which can be well- or poorly marginated. The detection of lesions $ \geq $ 3cm do not typically pose a diagnostic challenge \citep{samRefB}. However, detecting small pulmonary nodules on plain film is challenging, even despite high spatial resolution because an x-ray is a single projection of the entire 3D thorax volume. The planar nature of radiograph acquisition means that thoracic structures are superimposed, thus, the heart, diaphragm, and mediastinum may obscure a large portion of the lungs. Patients may also have several co-existing pathologies visible on each	 radiograph. Many benign lesions can mimic a pathology, due to composite shadowing, and, furthermore, the nodule can be very small or with ill-defined margins. Studies have shown that in up to 40\% of new lung cancer diagnoses, the lesion was present on previous plain film, but was missed and only picked up in hindsight \citep{forrest1981radiologic,samRefC}.
	
	Computer-aided detection (CAD) systems using machine learning techniques can facilitate the automated detection of lung nodules and provide a cost-effective {second-opinion reporting mechanism}. The reported performance of these CAD systems varies substantially depending on the size and nature of the samples. For instance, sensitivity rates reported in the literature for lesions larger than $5$mm vary from $51\%$-$71\%$  \citep{moore2011sensitivity, szucs2013comparison}. Currently, state-of-the-art results for automated object detection in images are obtained by deep convolutional neural networks (DCNN). During training, these methods require a large number of manually annotated images in which the contours of each object are identified or, at the very least, have a bounding box indicating their location within the image. The large majority of these methods use regression models to predict the coordinates of the bounding boxes \citep{Erhan2014, Szegedy13} or, alternatively, make use of sliding windows \citep{RenHG015, sermanet2014overfeat}. Most documented studies rely on large datasets of natural images ~\citep{Everingham2010,LinMBHPRDZ14} where the objects to be detected are typically well-defined and sufficiently within the context of the entire image. Fundamentally, the applicability of these technologies in radiology has not been fully explored, partially due to the paucity of large databases of annotated medical images.

	In recent years, {the increasing availability of digital archiving and reporting systems, such as PACS (Picture Archiving and Communication System) and RIS (Radiology Information System), is laying the technological foundations needed to make available large volumes of clinical data and images from hospital archives \citep{cho2015much, DBLP:journals/corr/CornegrutaBWM16}.}
	In this study, our aim is to leverage a large number of radiological exams extracted from a hospital's data archives to explore the feasibility of deep learning for lung nodule detection. In particular, we assess the performance of a statistical classifier that discriminates between chest radiographs with {elements/regions} indicating the presence of a pulmonary lesion and those that do not. Our first hypothesis is that, with a sufficiently large training database, a classifier based on deep convolutional networks can be trained to accomplish this task using only weak image labels. In order to address our hypothesis, we collected over $700,000$ historical chest radiographs from two large teaching hospitals in London (UK). A natural language processing (NLP) system was developed to parse all free-text radiological reports to identify all the exams containing pulmonary lesions. This is a challenging learning task as a proportion of automatically-extracted labels in the training dataset is expected to be erroneous or incomplete due to reporting errors or omissions (estimated to be at least $3$-$5\%$ \citep{Brady17}), inter-reader variability \citep{Elmore94,Elmore15} and potential NLP failures.  The performance of the resulting image classifier was assessed using a manually curated, independent dataset of over $6,000$ exams.
	
	Our second and main {hypothesis} is that significant classification improvements can be obtained by augmenting the weak and potentially noisy labels by using bounding boxes to indicate the exact location of any lesions in a subset of the training exams. Manual annotation  simply does not scale well given the size of currently available historical datasets; realistically only a fraction of all the exams can be reviewed and annotated. It would be, therefore, of interest to design a classifier that leverages both weakly labelled and annotated images. To investigate this hypothesis, approximately {$9\%$} of the radiographs presenting lesions were randomly selected and reviewed by a radiologist who manually delineated the bounding boxes. This annotation process resulted in over $3,000$ lesion examples. 
	
	We present two different learning strategies to leverage both weak labels and the annotations of lesions. Our guiding principle was that, when the position of a lesion is known during training, it can be exploited to provide the network with visual feedback that can inform on the quality of the features learned by the convolutional filters.  As such, both strategies introduce attention mechanisms within the classifier in order to learn improved imaging representations. Our first approach exploits a {\it soft attention mechanism}. Using weakly-labelled images, a convolutional network learns imaging features by minimising the classification error and generates saliency maps highlighting parts of an image that are likely to contain a lesion. Using the subset of annotated images, a composite loss function is employed to penalise the discrepancy between the network's implied position of a lesion, provided by the saliency map during training and the real position of the lesion. A large loss indicates that the network's current representation does not accurately capture the lesion's visual patterns, and provides an additional mechanism for self-improvement through back-propagation. The resulting architecture, a convolutional neural network with attention feedback (CONAF), features an improved localisation capability, which, in turn, {boosts} the classification performance. 
	
	\begin{figure*}[t!]
		\centering
		
		\vspace*{1mm}
		\begin{minipage}[b]{.30\textwidth}
			\centering
			\includegraphics[width=0.90\linewidth]{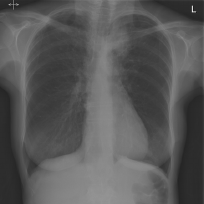}
			\captionsetup{labelformat=empty}
		\end{minipage}%
		\begin{minipage}[b]{.30\textwidth}
			\centering
			\includegraphics[width=0.90\linewidth]{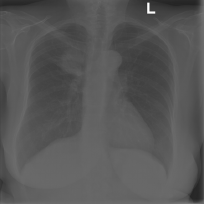}
			\captionsetup{labelformat=empty}
		\end{minipage}
		\begin{minipage}[b]{.30\textwidth}
			\centering
			\includegraphics[width=0.90\linewidth]{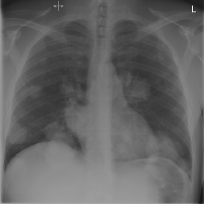}
			\captionsetup{labelformat=empty}
		\end{minipage}%
		
		\vspace{1mm}
		
		\begin{minipage}[b]{.30\textwidth}
			\centering
			\includegraphics[width=0.90\linewidth]{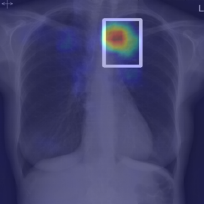}
			\captionsetup{labelformat=empty}
		\end{minipage}%
		\begin{minipage}[b]{.30\textwidth}
			\centering
			\includegraphics[width=0.90\linewidth]{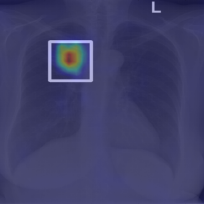}
			\captionsetup{labelformat=empty}
		\end{minipage}%
		\begin{minipage}[b]{.30\textwidth}
			\centering
			\includegraphics[width=0.90\linewidth]{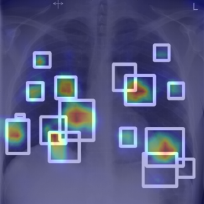}
			\captionsetup{labelformat=empty}
		\end{minipage}%

		\caption{Three examples (one per column) of successfully detected lung lesions on chest radiographs using CONAF {(convolutional neural network with attention feedback)}. The first row contains the original chest radiographs and the second one the probability heatmaps generated by CONAF along with the ground truth bounding boxes drawn in white by the radiologists. The heatmaps indicate the likely position of a lesion; high probability regions are in red and low probability regions are in blue.}
		\label{fig:results_intro}
	\end{figure*}
	
	Our second approach implements a {\it hard attention mechanism}, and specifically an extension of the Recurrent Attention Model (RAM)~\citep{BaMK14, Mnih14, SermanetFR14, Ypsilantis17}. In contrast to CONAF, each image is processed in a finite number of sequential steps. At each step, only at a portion of the image is used as input; each location is sampled from a probability distribution that leverages the knowledge acquired in the previous steps. The information accumulated through a random path image culminates in the classification of the image. The classification score acts as a reward signal which, in turn, updates the probability distribution controlling the sequence of image locations that should be visited. This results in more precise attention being paid to the most relevant parts of the image, i.e. the lungs. Our proposed architecture, RAMAF (Recurrent Attention Model with Attention Feedback), rewards a higher classification score when the {\it glimpses} attended by the algorithms during training overlap with the correct lesion locations. Establishing this improves the rate of learning, yielding a faster convergence rate and increased classification performance.  
	
	The article is structured as follows. In Section~\ref{sec:data}, we introduce the dataset used in our experiments and explain how the chest radiographs have been automatically labelled using a natural language processing system. The CONAF and RAMAF algorithms are presented in Sections~\ref{sec:firstmodel} and \ref{sec:secondmodel}, respectively. Their performance has been assessed and compared to a number of alternative architectures that use either weak labels or annotated images. In Section \ref{sec:results}, we describe our experimental results supporting the hypothesis that leveraging a relatively small portion of manually annotated lesions, in addition to a large sample of weakly-annotated training examples, can drastically enhance the classification performance.
	
	\section{A repository of chest radiographs} \label{sec:data}
	
	For this study, we obtained a dataset consisting of $745,479$ chest x-ray exams collected from the historical archives of Guy's and St. Thomas' NHS Foundation Trust in London from January $2005$ to March $2016$.  For each exam, the free-text radiologist report was extracted from the RIS (Radiology Information System). For a subset of  $634,781$ exams, we were also able to obtain the DICOM files containing pixel data. All paediatric exams ($ \leq 16$ years of age) were removed from the dataset resulting in a total of $430,067$ exams for which {both images and radiological reports were available}. DICOM headers and any reports were data anonymised and an ethics committee waiver of consent was granted for the study. {The size of original radiographs ranged between $734 \times 734$ to $4400 \times 4400$ pixels}, so we scaled each image to a standard size of $448 \times 448$ to keep the computational requirements to a sustainable level, but otherwise no other preprocessing was carried out. 
	
	The radiological reports were used as the ground truth to determinate whether any chest radiograph in our database contained evidence of a suspected lung lesion. For this study, we tagged each exam using three mutually exclusive labels: (a) {\it normal}, i.e. exams presenting no radiological abnormalities; (b) {\it lesions}, i.e. exams reported as presenting at least a focal lesion; (c) {\it others}, i.e. exams that are not normal, but do not contain a pulmonary lesion.  The labelling task was automated by using an extension of an NLP system originally developed for the detection of clinical findings from radiological reports \citep{DBLP:journals/corr/CornegrutaBWM16}; an overview of the NLP system and its associated validation study can be found in \ref{appendix:nlp}. The NLP system identified $101,766$ normal exams, $23,132$ exams containing at least one lesion and $305,169$ exams having radiological abnormalities, other than suspected lesions. 
	
\begin{figure}[t!]
		\centering
		\includegraphics[scale=0.30]{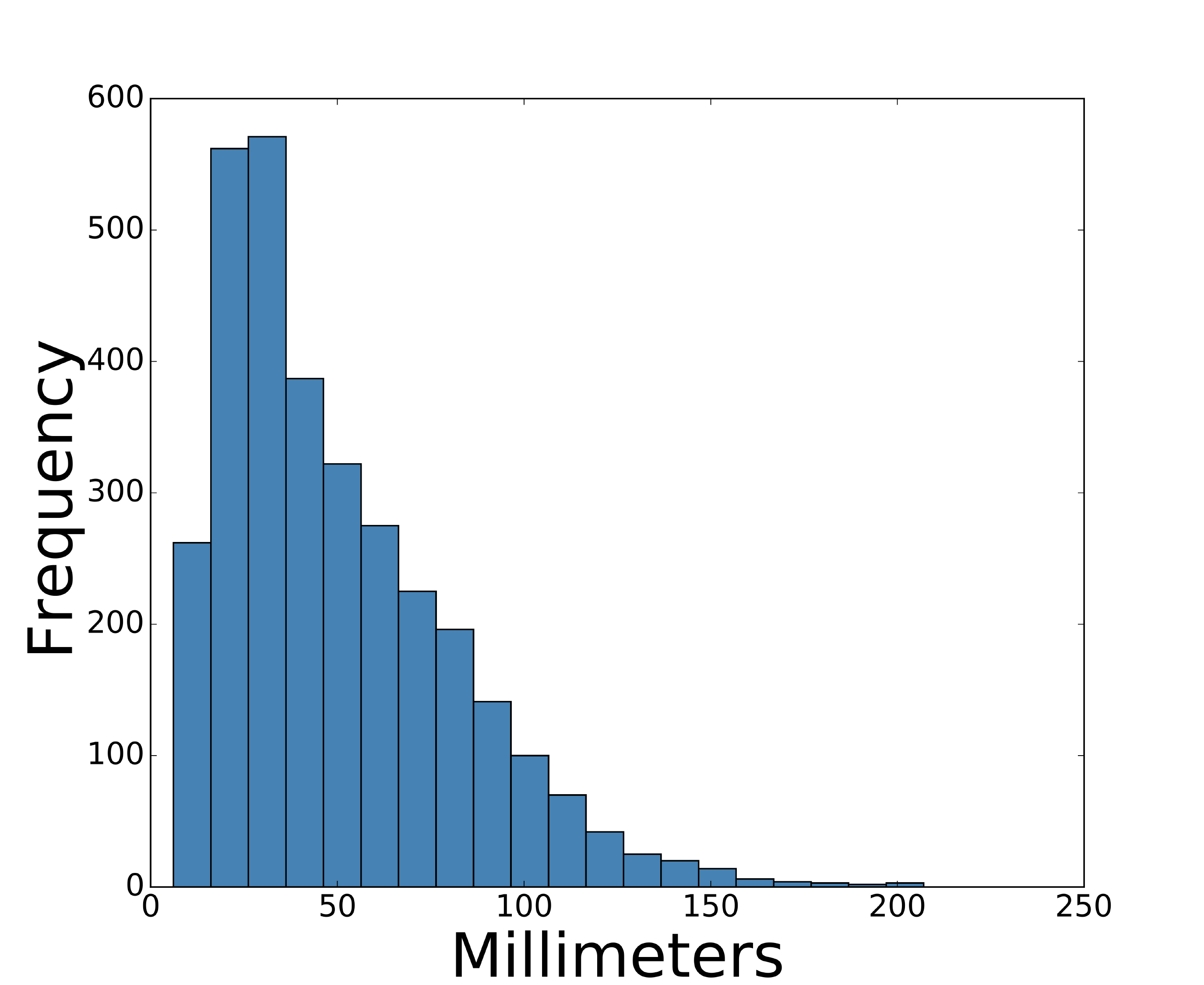}
		\caption{Frequency distribution of lesion size across all the annotated images; the size is measured in millimeters and represents the maximal width of the bounding box.}
		\label{fig:hist_bb}
	\end{figure}

	The most common appearances of a pulmonary nodule is that of a small, rounded opacity within the lung. However, lesions can be solid, semi-solid or groundglass; well- or ill-defined; single or multiple; and can occur anywhere in the lung. On radiograph, they can overlap with the ribs, the mediastinum, the diaphragm, or the heart. According to accepted nomenclature, a nodule is $<3$cm; a mass is $\geq 3$cm, although for this study we have used the term \emph{lesion} to include both criteria \citep{samRefB}. With the aim of improving the sensitivity of the detection of lung tumours on chest x-rays, the CAD system was trained and then tested on radiographs that reported a possible pulmonary lesion, not just those cases with CT (computed tomography) or {hystopathological confirmation} of cancer. Amongst all the $23,132$ {images} containing lesions in our database, $2,196$ were manually annotated by an experienced radiologist resulting in $ 3,253 $ annotated lesions (see also Section \ref{sec:noduleclassificationperfomance}).  A bounding box was drawn around each lesion within each image; see Fig. \ref{fig:results_intro} for some examples. The approximate size of a lesion was measured by taking the longest side of the bounding box in millimeters. This measurement only provides an upper bound of the real nodule's size; Fig. \ref{fig:hist_bb} shows the frequency distribution of lesions {diameters}.

	\section{Proposed architectures}
	
	\subsection{Convolution networks with attention feedback (CONAF)} \label{sec:firstmodel}
	
	In this section we set out our proposal of an image classifier based on deep convolutional neural networks. Our aim is to detect chest radiographs that are likely to contain one or more lesions. Although the localisation of the lesions within an image is not our primary interest, this information can be extracted from a trained network to generate saliency maps, i.e. heatmaps indicating where the lesions are more likely to be located within the original x-ray.  Our proposed architecture exploits these maps to introduce a soft attention mechanism. For radiographs containing annotations, the saliency maps can be compared against the ground truth (the bounding box drawn by radiologist) to derive a localisation error. Although this additional error term can only be computed for a subset of images during training time, it provides useful feedback about the most likely inferred position of a lesion at any given time, and this information can be leveraged to further decrease the classification error.
	
	All the available radiographs are collected in a set $\mathcal{X}^{(w)} = \{\textbf{x}_i \in \mathbb{R}^{448\times448}; i=1, \ldots, N_w \}$ with corresponding labels in $\mathcal{Y}^{(w)} = \{ y_i \in \{0,1,2\}; i=1, \ldots, N_w \}$. In our dataset, $N_w=430,067$. A label $y_i=0$ indicates that the exam has been reported as normal (i.e. there are no radiological abnormalities) whereas $y_i=1$ indicates the presence of one or more lesions and $y_i=2$ refers to other reported abnormalities other than pulmonary lesions. {All the images which contain lesions that have been annotated with bounding boxes are collected in a subset}  $\mathcal{X}^{(b)} \subset \mathcal{X}^{(w)}$, which has cardinality $N_b < N_w$. In our dataset, $N_b=2,196$. The corresponding bounding box annotations are collected in a set of binary masks, $\mathcal{B}^{(b)} =  \{ \mathbf{b}_i \in \{ 0,1 \}^{448 \times 448}; i=1, \ldots, N_b \}$ with ones indicating pixels belonging to a lesion and zeros being background pixels.   
	
\begin{figure*}[t!]
		\centering
		\includegraphics[scale=0.60]{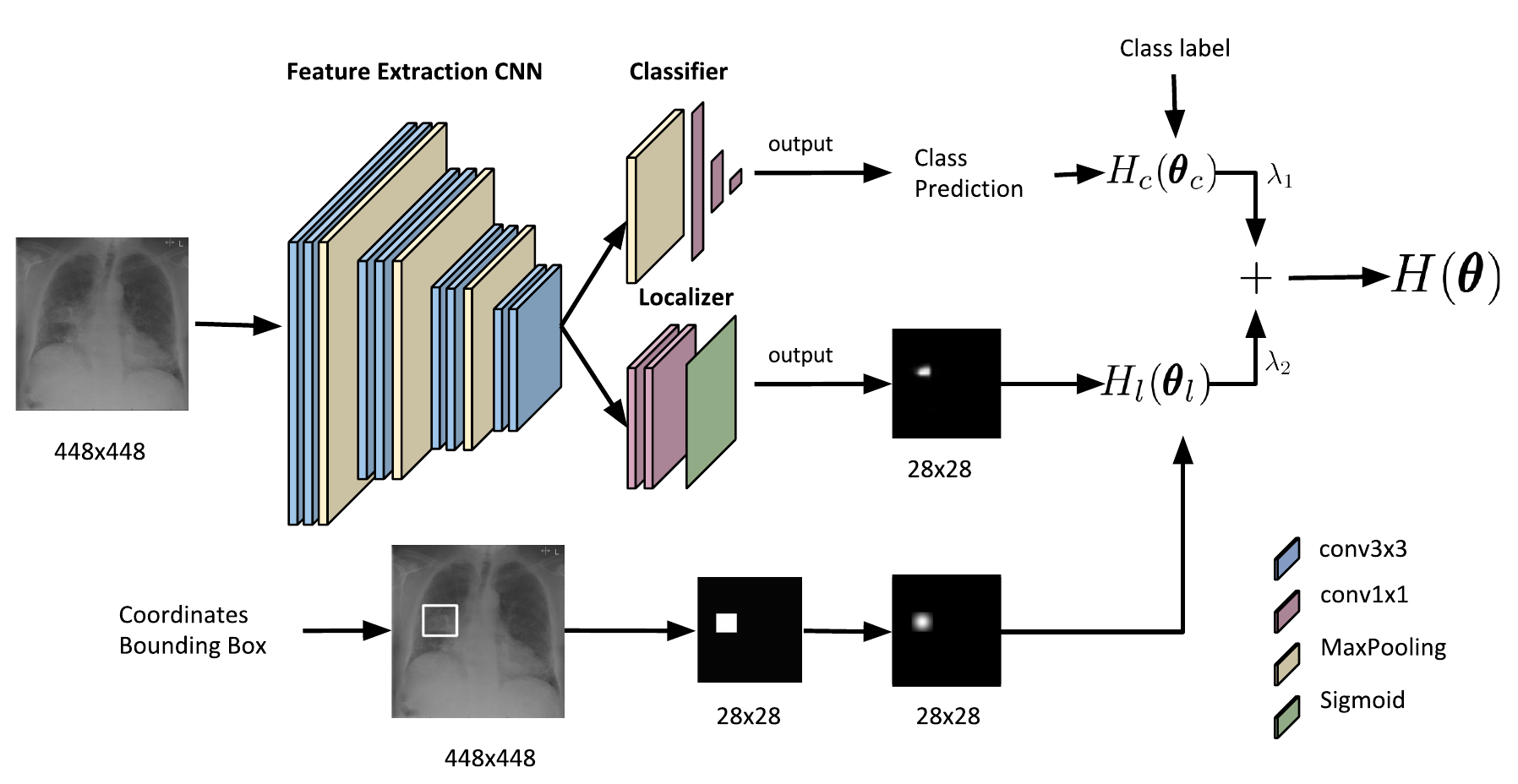}
		\caption{An illustration of the CONAF model. Both the classifier and localizer receive as input the output of the Feature Extraction CNN. The localisation loss function $H_{l}$ and the classification loss function $H_{c}$ are linearly combined to form the hybrid loss function $H$.}
		\label{fig:arch1}
	\end{figure*}

	Our proposed architecture is presented in Fig. \ref{fig:arch1}. It relies upon three building blocks: a convolutional neural network for feature extraction and two separate components used for classification and localisation. The feature extraction block takes $\textbf{x}_i$ as input and consists of a sequence of convolutional layers and max-pooling layers. Our implementation here is similar to the commonly used VGG13 \citep{simonyan2014very}, which has proven reliable in our studies. 
	{The last layer terminates with $512 \times28\times28$ feature maps which contain high-level representation of the input image and are used as inputs for both the classification and localisation components. After performing a global max pooling operation which outputs $512$ units, the classification branch consists of two layers ($512 \mapsto 256 $ and $ 256 \mapsto 2$ ) of $1 \times 1$ convolutions (equivalent to fully connected layers) inferring the probability that the input image is assigned to a class.}
	We considered two different binary classification problems: a simplified one, where those x-rays presenting with lesions are compared to those without any radiological abnormalities, i.e. {\it Lesion vs. Normal}, and a realistic one, {\it Lesion} vs. everything else (i.e. the union of {\it Normal} and {\it Others}). The latter is significantly more challenging as the {\it Others} class contains a very large number of radiological abnormalities, some of which often co-exist with the lesions observed in the {\it Lesion} class.
	
	The input for this branch consists of all images in $\mathcal{X}^{(w)}$. All shared weights for feature extraction and the weights which are specific to the classification branch are collected in a parameter vector $\pmb{\theta}_c$, which is optimised by minimising the binary cross-entropy loss, 
	\begin{equation*} \label{eq:hc}
	H_c(\pmb{\theta}_c) = - \frac{1}{N_w}\sum\limits_{i=1}^{N_w}[y_i \log(\hat{y}_i)+(1-y_i) \log(1-\hat{y}_i)]
	\end{equation*}
	where $\hat{y}_i $ is the predicted class. 
	
	The images in $\mathcal{X}^{(b)}$ contribute towards a second loss term, which is computed by the localisation component consisting of a series of two layers performing $1 \times 1$ convolutions {($512\times28\times28 \mapsto 256\times28\times28$ and $256\times28\times28 \mapsto 1\times28\times28$)}. The output is passed through a sigmoid function to produce a scoremap $\phi(\textbf{x}_i) \in [0,1]^{28\times28}$ used to infer the position of lesions within the image. Values away from zero and closer to one indicate that the corresponding pixels are likely to contain a lesion. Our rationale here consists of comparing a scoremap with the associated ground-truth binary mask, $\textbf{b}_i$ in order to quantify the current localisation error. An adjustment step is required at this stage since the manually delineated masks are rectangular or {square} in shape whilst the true lesions are generally round. Since all manually annotated lesions are typically {centred} in the middle of the bounding box, we {use a 2D Gaussian kernel} to trace an elliptical area of high probability in the middle of the box,
	\begin{equation}
	\mathcal{G}(r_1,r_2) = \frac{1}{2\pi\sigma^2} e^{- \frac{\frac{r_1}{2}^2+\frac{r_2}{2}^2}{2\sigma^2}},
	\end{equation}
	where $ r_1 $ and $ r_2 $ are the length and width, respectively, of the bounding box and $\sigma$ controls the size of the lesion within the box. We then resize the original mask to obtain $ \textbf{z}_i \in [0,1]^{28\times28} $, which is now directly comparable to $ \phi(\textbf{x}_i) $. A pixel-wise mean-square loss is then computed as $\textbf{e}_i = \lVert \phi^*(\textbf{x}_i) - \textbf{z}_i \rVert^2$, where
	\begin{equation*}
	\phi^*(\textbf{x}_i) = \frac{\phi(\textbf{x}_i)}{\max{\phi(\textbf{x}_i)}} 
	\end{equation*}
	is a rescaled normalised scoremap. The proposed scaling {ensures that the prediction values are in range $[0,1]$, so to be properly compared to $\textbf{z}_i$}.
	The final localisation loss is defined as 
	\begin{equation*}\label{eq:hl}
	H_l(\pmb{\theta}_l) = \frac{1}{{N_b}} \sum\limits_{i=1}^{N_b} \Biggl\lVert \frac{ \textbf{e}_i }{ \alpha - \textbf{z}_i} \Biggl\lVert^2
	\end{equation*}
	where $ \pmb{\theta}_l $ denotes all the network's weights and the sum is over all images containing a bounding box. 
	Given that lung lesions cover only a small part of the image, we expect only a minority of pixels to contribute to the above error. The loss term above places more importance to high-value pixels by diving each $\textbf{e}_i $ by $\alpha - \textbf{z}_i $, where $ \alpha $  is a constant set to $1.1$; see also \citet{cornia2016deep}.  The overall network architecture in Fig. {\ref{fig:arch1}} is then trained end-to-end as to minimise a linear combination $ H(\pmb{\theta}) $ of classification and localisation losses, i.e. 
	\begin{equation*}
	H(\pmb{\theta}) = \lambda_1 H_c(\pmb{\theta}_c) + \lambda_2 H_l(\pmb{\theta}_l),
	\end{equation*}
	where $\lambda_1$ and $\lambda_2 $ are positive scalars controlling the trade-off between the two errors. Further implementation details are provided in Section \ref{sec:results}.
	
	\begin{figure*}[t!]
		\begin{center}
			\includegraphics[scale=0.25]{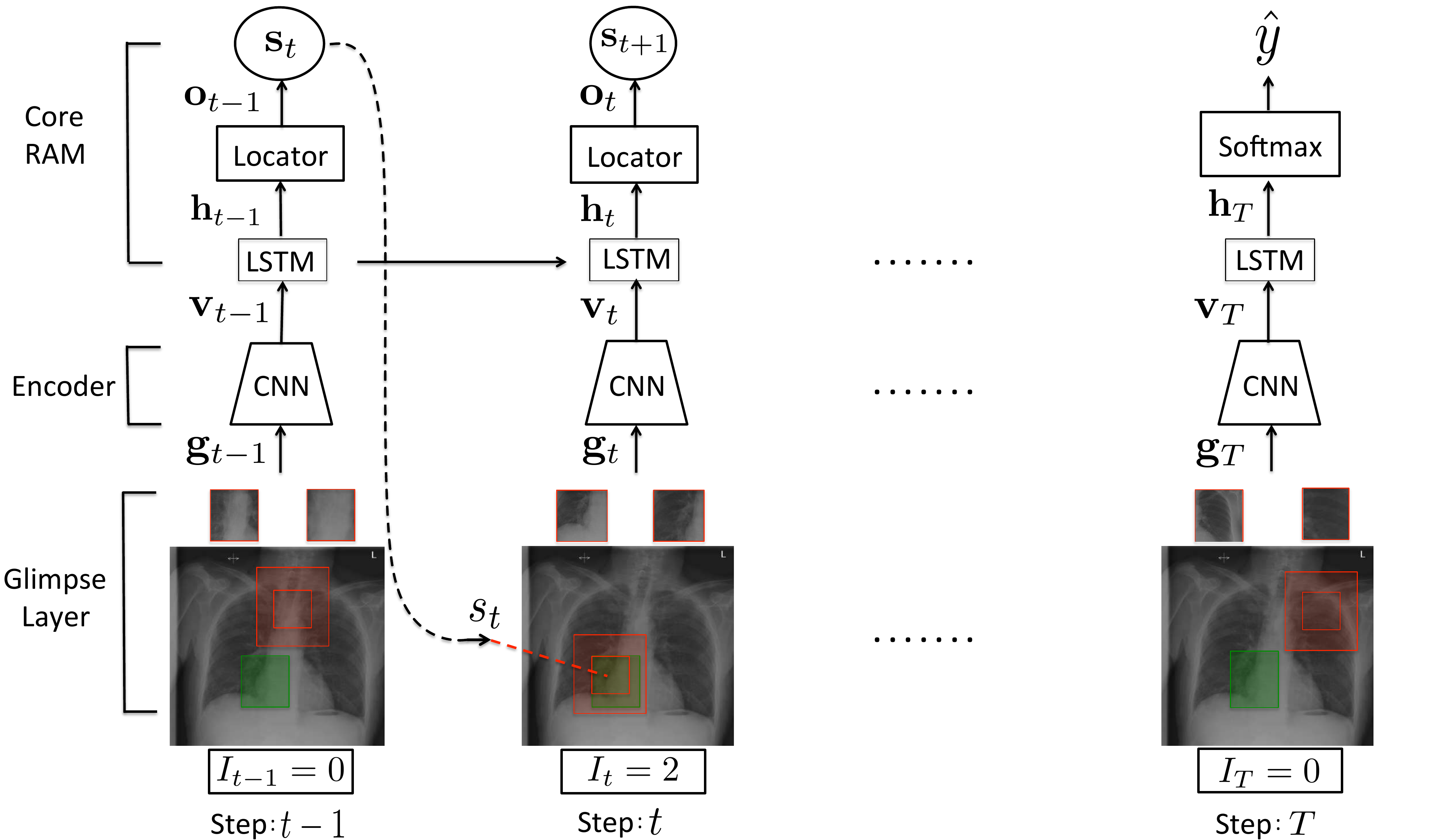}
		\end{center}
		\caption{An illustration of the RAMAF model. The green colour frame represents the bounding box annotation and red colour frames represent the proposed \enquote{glimpses} at each time step $t$. At each time step $t$ the Core RAM samples a location $\textbf{s}_{t}$ of where to attend next. In time steps (see time step: $t$) where the model samples a location that belongs to the bounding box annotation, it receives an extra reward.}
		\label{fig:ram}
	\end{figure*}
	
	\subsection{Recurrent attention model with annotation feedback (RAMAF)} \label{sec:secondmodel}
	
	Here, we propose an extension of the original RAM model \citep{Mnih14}, which we call Recurrent Attention Model with Annotation Feedback (RAMAF). The model works by observing only one portion of the image, one glimpse, at a time, and learns to navigate through an image by taking a sequence of glimpses at strategically chosen positions. After each step, the algorithm has observed a larger portion of the overall image, and the optimal policy controlling \enquote{where to look} minimises the classification error. In what follows, $\textbf{g}_{i,t}$ represents the observation, i.e. the glimpse seen by the model at time step $t$, and $\textbf{s}_{i,t} \in \mathbb{R}^{2}$ represents the coordinates $(\texttt{x}_{i,t}, \texttt{y}_{i,t})$ of the pixel located at the centre of the glimpse. The overall sequence of glimpses seen by the model for an image $\textbf{x}_{i}$ is defined as $\textbf{S}_{i,1:T} = \{\textbf{s}_{i,1}, \textbf{g}_{i,1}, \textbf{s}_{i,2}, \textbf{g}_{i,2}, \hdots, \textbf{s}_{i,T}, \textbf{g}_{i,T}\}$. In our formulation, each glimpse consists of two image patches of different size sharing the same central location $\textbf{s}_{t}$, each one capturing a different context around the same region. The largest patch is scaled down to match the size of the smallest one (see Fig.~\ref{fig:ram}). Once $\textbf{S}_{i,1:T} $ is available, a reward signal is generated depending on whether the image has been correctly classified. In RAMAF, in addition to this classification reward, an additional reward signal is introduced to take into account the number of central coordinates $\textbf{s}_{i,t}$ that lie within the coordinates of the bounding boxes for the images in $\mathcal{X}^{(b)}$.     
	
	Fig.~\ref{fig:ram} provides an overview of the model. On top of the glimpse layer, an encoder is introduced to compress the information contained in the glimpse and extract a representation that is robust to noise. The encoder implemented here differs from the one used originally in \citep{Mnih14}. In this application, we have a complex visual environment featuring high variability in both luminance and object complexity. This is due to the large variability in a patient's anatomy, as well as technical variability, since the radiographs in our dataset were acquired from over $40$ different x-ray devices. At this stage, each glimpse is passed through a stack of two convolutional layers followed by max-pooling operations. Each convolutional layer in the stack is pre-trained offline {on the training data} using convolutional auto-encoders with max-pooling \citep{Masci11} and then fine-tuned as part of end-to-end training for the RAMAF model. During training, each $\textbf{g}_{i,t}$ is concatenated with the location representation and passed as input to a fully connected layer, whose output is denoted as $\textbf{v}_t \in \mathbb{R}^{256}$. The output is then passed as input to the Core RAM model, as illustrated in Fig.~\ref{fig:ram}. 
	
	The role of the Core RAM model is to summarise the information extracted from the sequence of glimpses and use this summary to decide where to attend next. In our formulation, the information summary is formed by the hidden representation $\textbf{h}_{t} \in \mathbb{R}^{256}$ of a recurrent neural network with long short-term memory (LSTM) units. At each time step $t$, the encoder's output vector $\textbf{v}_{t}$ and the previous hidden representation $\textbf{h}_{t-1} \in \mathbb{R}^{256}$ of the RNN are passed as input to the current LSTM unit. The Locator (see Fig.\ref{fig:ram}) receives the hidden representation $\textbf{h}_{t}$ from the LSTM unit and passes on to a fully connected (FC) layer, resulting in a vector $\textbf{o}_{t} \in \mathbb{R}^{2}$ (see Fig.~\ref{fig:ram}). The Locator decides the position of the next glimpse by sampling $\textbf{s}_{t+1} \sim N(\textbf{o}_{t}, {\bf \Sigma})$, i.e. from a normal distribution with mean $\textbf{o}_{t}$ and diagonal covariance matrix ${\bf \Sigma}$. At the very first step, we initiate the algorithm at the centre of the image, and always use a fixed covariance matrix, ${\bf \Sigma}$.
	
	For each $\textbf{x}_{i} \in \mathcal{X}^{(b)}$, we use a spatial reward function that takes advantage of the bounding box annotations, i.e.
	$$
	R(\textbf{S}_{i,1:T}) = r_{i} + \frac{1}{T}\sum_{t=1}^{T}I_{t}
	$$
	consisting of two components. First, $r_{i} =1$ if the image classification is correct, otherwise $r_{i}= 0$. We set $I_{t}=2$ if the glimpse's central pixel $\textbf{s}_{t}$ at time step $t$ lies within the annotation bounding box, and $I_{t}=0$ otherwise (see Fig. \ref{fig:ram}). The latter term represents a spatial reward signal, which needs to be minimised. The model is then trained to learn a policy that maximises the conditional probability of the true label given the partial interaction with the radiographs. As in \citet{Mnih14}, we optimise the cross-entropy loss to train the network to correctly classify the radiographs. We train the part of the model which proposes the observation locations using the REINFORCE algorithm; further details can be found in \ref{appendix:ramaf}.  
	
	\section{Experimental results} \label{sec:results}
	
	\subsection{Further implementation details}
	
	In this section we provide additional implementation details.  The CONAF loss function was fully specified using $\lambda_1 = 10 $ and $\lambda_2 = 0.1 $ as these parameters yielded optimal performance on the validation test.
	Training was done using back-propagation with adadelta \citep{zeiler2012adadelta}, mini-batches of 32 images and a learning rate of 0.03.  
	During the training we fed the network through two types of mini-batches: one is composed {by only} images associated to weak labels and the other is composed of images with bounding box annotations. We picked the former type with probability $ p=0.8 $ and the latter with $ p=0.2 $. This approach was followed to avoid over-fitting in the localisation part since the number of annotated images was significantly smaller than the overall number of images.
	Given the unbalanced sample sizes characterising our dataset, all the images within a mini-batch were randomly selected ensuring that half of them were labelled as \emph{Lesion} and other half as either \emph{Normal} or \emph{Others}, depending on the experiment. 
	{The $\sigma$ parameter controlling the 2D Gaussian Kernel was set to $ 0.25 $ in order to draw elliptical areas in which values close to zero are not too far or too close to the bounding box edges. A narrow Gaussian kernel would not allow using all the information available in the annotation, while a loose one would lead to a loss of precision. This has been done to provide a better approximation of the lesion shapes, since usually nodules and masses are round-shaped, while our annotations are rectangular.}
	
	For the RAMAF model, we used a fixed length of $7$ glimpses, each one containing a high-resolution window of size $70\times70$ pixels and a low-resolution window of size $140 \times 140$ pixels. The convolutional layers within the encoder consisted of $16$ feature maps with filters of dimension $3 \times 3$. These were followed by max-pooling layers with a non-overlapping receptive field of dimension $2 \times 2$. For training the model we used back-propagation through time (BPTT)~\citep{Werbos90}, optimized with Adam~\citep{KingmaB14} with mini-batches of size $40$ and learning rate of $0.0001$. The number of annotated images within each mini-batch varied between $5$ and $20$. The weights of the Core RAM were initialized with randomly selected values from a uniform distribution over the interval $[-0.1,0.1]$. A diagonal covariance matrix ${\bf \Sigma}$ with elements $0.22$ was used for sampling each glimpse's coordinates.

	\subsection{Competing architectures}
	
	Other neural network architectures were tested in comparison to our algorithms. To assess the degree of both the classification and localisation performance achievable from using weak labels only, we used two state-of-the-art weakly-supervised methods performing both classification and localisation tasks. The first method, proposed in \citet{oquab2015object}, uses convolutional adaptation layers at the end of the feature extraction layer in order to get a scoremap for each class. The second method, proposed in \citet{zhou2015learning}, uses a global average pooling layer, after the last layer of feature maps in order, to encourage the network to identify the complete extent of the object; it then passes the output features as inputs to a fully connected layer in order to compute the desired output. Saliency maps are obtained by projecting back the weights of the fully connected layer on to the last layer of convolutional feature maps. 
	
	Furthermore, we considered two state-of-the-art fully supervised methods for object detection. The OverFeat algorithm performs classification, localisation and detection \citep{sermanet2014overfeat}. It scans an image in a sliding window fashion at several scales; during training the tasks of classification and bounding box prediction are performed simultaneously. In a final stage, all predicted bounding boxes are merged according to a proposed scheme. The second algorithm uses a CNN module to encode an image in high-level feature representation, which is then passed to a LSTM (long-short term memory) network which learns to decode this representation into predicted bounding boxes \citep{Stewart2015End}.
	
	\begin{table}[t!]
		\centering 
		\caption{SUMMARY OF AVAILABLE SAMPLE SIZES.}
		\begin{tabular}{|c|c|c|c|c|}
			\hline  Radiological appearance & Train & Validation & Test & Total \\ 
			\hline  Normal  & 88,929 & 11,118 & 1,719 & 101,766 \\    
			\hline  Lesion  & 18,870 & 2,398 & 1,864 & 23,132  \\
			\hline  Others & 267,326 & 33,576 & 4,267 & 305,169  \\ 
			\hline 
		\end{tabular}		 
		\label{table:splits}
	\end{table}
	
	\begin{table*}[!t]
		\caption{CLASSIFICATION PERFORMANCE: LESION VS NORMAL ONLY AND LESION VS ALL OTHERS.}
		\label{table:normal_vs_all}
		\centering
		\resizebox{\columnwidth}{!}{%
			\begin{tabular}{|c|c|c|c|c||c|c|c|c|}
				\hline   & \multicolumn{4}{c||}{Lesion vs Normal Only} & \multicolumn{4}{c|}{Lesion vs All Others} \\ 
				\hline  Method 				& Accuracy  & F1  & Sensitivity & Precision & Accuracy  & F1  & Sensitivity & Precision \\ 
				\hline 
				\hline  OverFeat \citep{sermanet2014overfeat} & 0.75 & 0.76 & 0.77 & 0.75 & 0.64	 & 0.55	 & \textbf{0.77} & 0.42  \\ 
				\hline  Stewart  \citep{Stewart2015End} & 0.75 & 0.74 & 0.72 & 0.78 & 0.64	 & 0.54	 & 0.73 & 0.43  \\    
				\hline  Oquab \citep{oquab2015object}  & 0.81 & 0.79 & 0.72 & 0.89  & 0.61 & 0.46 & 0.44 & 0.48  \\ 
				\hline  Zhou \citep{zhou2015learning}  & 0.81  & 0.79 &  0.71 & 0.89 & 0.72  & 0.62 &  0.53 & \textbf{ 0.74} \\
				\hline  RAM & 0.72 & 0.70  & 0.63 & 0.78 & 0.61 & 0.48  & 0.52 & 0.44  \\\Xhline{2\arrayrulewidth}
				\hline  RAMAF & 0.73 & 0.74 & 0.74 & 0.74  & 0.61 & 0.47 &  0.52 &  0.43 \\  
				\hline  CONAF & \textbf{0.85} & \textbf{0.85}  & \textbf{0.78} & \textbf{0.92} & \textbf{0.76} & \textbf{0.67}  & 0.74  &  0.60 \\
				\hline 
			\end{tabular}		 
		}
	\end{table*}
	
	\begin{table*}[!t]
		\centering
		\caption{LOCALISATION PERFORMANCES: LESION VS NORMAL ONLY AND LESION VS ALL OTHERS.}
		\label{table:locresult}
		\resizebox{\columnwidth}{!}{%
			\begin{tabular}{|c|c|c|c||c|c|c|c|c|c|}
				\hline   & \multicolumn{3}{c||}{Lesion vs Normal Only} & \multicolumn{3}{c|}{Lesion vs All Others} \\ 
				\hline  Method 				& Sensitivity   	& Precision  	&  Average Overlap  & Sensitivity   	& Precision  	&  Average Overlap \\
				\hline  OverFeat \citep{sermanet2014overfeat}  & 0.35 & 0.41 & 0.27 & 0.37	& \textbf{0.28} & 0.30	\\ 
				\hline  Stewart  \citep{Stewart2015End}  & 0.36 & \textbf{0.47} & 0.26 & 0.37 & 	\textbf{0.28} & 0.30 \\ 
				\hline  Oquab \citep{oquab2015object} & 0.57 & 0.14 	& 0.22 & 0.02 & 0.01  & 0.02 \\
				\hline  Zhou \citep{zhou2015learning} & 0.49 & 0.12 	& 0.25	& 0.34 & 0.10 	& 0.17	\\ \Xhline{2\arrayrulewidth}
				\hline  CONAF 	& \textbf{0.74} & 0.21  & \textbf{0.45}	& \textbf{0.65} & 0.15 	&	\textbf{0.43} \\ 
				\hline 
			\end{tabular} 	
		} 
	\end{table*}
	
	\subsection{Lesion classification performance}\label{sec:noduleclassificationperfomance}
	
	Comparison with these state-of-the-art methods for classification and localisation were conducted in two separate experiments. In the first experiment (\emph{Lesion vs Normal}), we assess the ability of our proposed models to differentiate between chest radiographs with normal radiological appearance (i.e. no abnormal findings) and chest radiographs with lesions. In the second experiment (\emph{Lesion} vs everything else), we tested whether our models were able to differentiate between chest radiographs with lesions and all other chest radiographs, including normals and those with other radiological findings ({\it Normal + Others}) (see Tables~\ref{table:normal_vs_all} and~\ref{table:locresult}). In both cases, we split the dataset into training (80\%), validation (10\%) and from the remaining set (10\%) we extracted our test set composed of $ 6,131 $ images which had had weak labels manually validated by two independent radiologists. This is necessary so as to ensure that our test set does not contain NLP errors. Each set is generated by randomly sampling from all available exams and ensuring that the patient's age and all the pathologies (see Table \ref{tab:NLP_results}) are represented. Furthermore, the training set contains the 80\% of images annotated with bounding boxes, whilst the test set contains the 20\%. Indeed we chose our model taking the classification F1 score on the validation set in order, to allow more bounding boxes examples for evaluating the localisation performance. All performance metrics reported here were calculated using the independent test set only. While the positive class (\emph{Lesion}) has been fixed, the negative class can vary between \emph{Normal} and \emph{Others},  according to the experiment we considered. 
	Table \ref{table:splits} provides the sample sizes. 
	For this task, we report on average accuracy, F1 measure, sensitivity and precision (see Table~\ref{table:normal_vs_all}).  
	We observe that CONAF outperforms all others methods in terms of average accuracy, F1-measure and sensitivity while the highest precision for the detection of images with lesions (vs Others) is achieved by the method using Class Activation Maps \citep{zhou2015learning}. It should be noted that, in this application, achieving the highest possible sensitivity is critical as the main aim is to minimise the percentage of possible tumours that are missed by the algorithm. 
	The accuracy of CONAF with respect to lesion size is illustrated in Fig. \ref{fig:results_3}. 
	{We calculated the deciles of the lesion size distribution (Figure \ref{fig:hist_bb}) to show the performances of both experiments. In both cases the accuracy increases linearly with the lesion size: nodules with a diameter smaller than $10$ millimeters are detected with an accuracy minor of $0.2$, while masses with a diameter bigger of $100$ millimeters are detected with an accuracy minor of $0.7$. This is an expected result since smaller nodules are more difficult to spot while largest masses are easy to detect for both humans and AI.}
	Furthermore, it can be noticed that RAMAF achieves better performance compared to the simpler RAM model trained without bounding boxes. 
	Both models are, in general, comparable to competing architectures in terms of overall performance. These results provide evidence that deep learning algorithms trained on a sufficiently large dataset are robust against a moderate level of label noise, which corresponds with findings from previously reported studies  \citep{2017arXiv170501936N, Rolnick17, DBLP:journals/corr/abs-1711-00583}.
	
	\subsection{Lesion localisation performance}
	
	The localisation performance was assessed by first segmenting the lesions against the background in the dataset. This was done by using the inferred scoremaps $ \phi(\textbf{x}_i) $  provided by CONAF and selecting all pixels whose estimated values on the maps were below a given threshold. We tried different threshold values ranging from $0.2$ to $0.8$ in increments of $0.2$. 
	After the thresholding process, we considered as lesion candidates the resulting regions with spatially contiguous pixels. A bounding box was drawn around each of these candidates. Any candidate bounding box that overlapped by at least $25\%$ with the ground truth bounding box was taken as a true positive. The number of true positive, false negative and false positive boxes was used to derive precision and sensitivity measures; that said, while it was not possible to calculate average accuracy and the F1 measure since that there are no true negatives for this task.
	
\begin{figure*}[h]
		\centering
		
		\vspace*{1mm}
		\begin{minipage}[b]{.25\textwidth}
			\centering
			\includegraphics[width=0.75\linewidth]{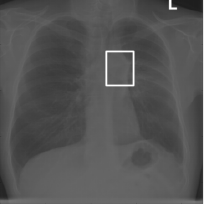}
			\captionsetup{labelformat=empty}
		\end{minipage}%
		\begin{minipage}[b]{.25\textwidth}
			\centering
			\includegraphics[width=0.75\linewidth]{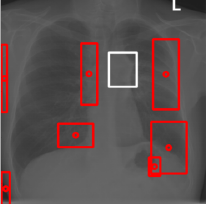}
			\captionsetup{labelformat=empty}
		\end{minipage}%
		\begin{minipage}[b]{.25\textwidth}
			\centering
			\includegraphics[width=0.75\linewidth]{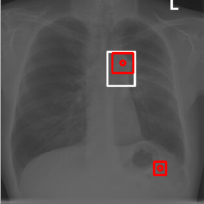}
			\captionsetup{labelformat=empty}
		\end{minipage}%
		\begin{minipage}[b]{.25\textwidth}
			\centering
			\includegraphics[width=0.75\linewidth]{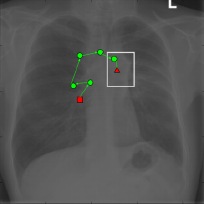}
			\captionsetup{labelformat=empty}
		\end{minipage}%

		\vspace{1mm}
		
		\begin{minipage}[b]{.25\textwidth}
			\centering
			\includegraphics[width=0.75\linewidth]{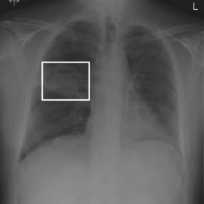}
			\captionsetup{labelformat=empty}
			\caption*{Original Image}
		\end{minipage}%
		\begin{minipage}[b]{.25\textwidth}
			\centering
			\includegraphics[width=0.75\linewidth]{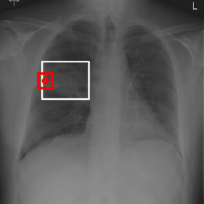}
			\captionsetup{labelformat=empty}
			\caption*{\citep{zhou2015learning}}
		\end{minipage}%
		\begin{minipage}[b]{.25\textwidth}
			\centering
			\includegraphics[width=0.75\linewidth]{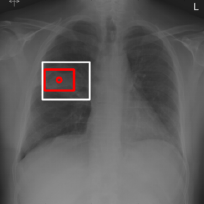}
			\captionsetup{labelformat=empty}
			\caption*{CONAF}
		\end{minipage}%
		\begin{minipage}[b]{.25\textwidth}
			\centering
			\includegraphics[width=0.75\linewidth]{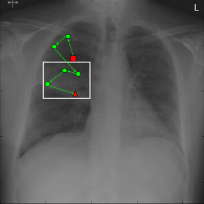}
			\captionsetup{labelformat=empty}
			\caption*{RAMAF}
		\end{minipage}%
		
		\caption{Two examples (one per row) of lesion localisation performance using different neural networks. The white boxes were manually drawn by radiologists. The red boxes are those considered likely to contain a lesion by the architecture described in \citep{zhou2015learning} and CONAF, , including false positives. For RAMAF, we display the trajectories followed by the algorithms before making a classification decision: the path starts at the point indicated by the red square and ends at at the point indicated by the red triangle.}
		\label{fig:results_1}
	\end{figure*}

\begin{figure*}[t!]
		\centering	
		\begin{minipage}[b]{.49\textwidth}
			\includegraphics[width=1\linewidth]{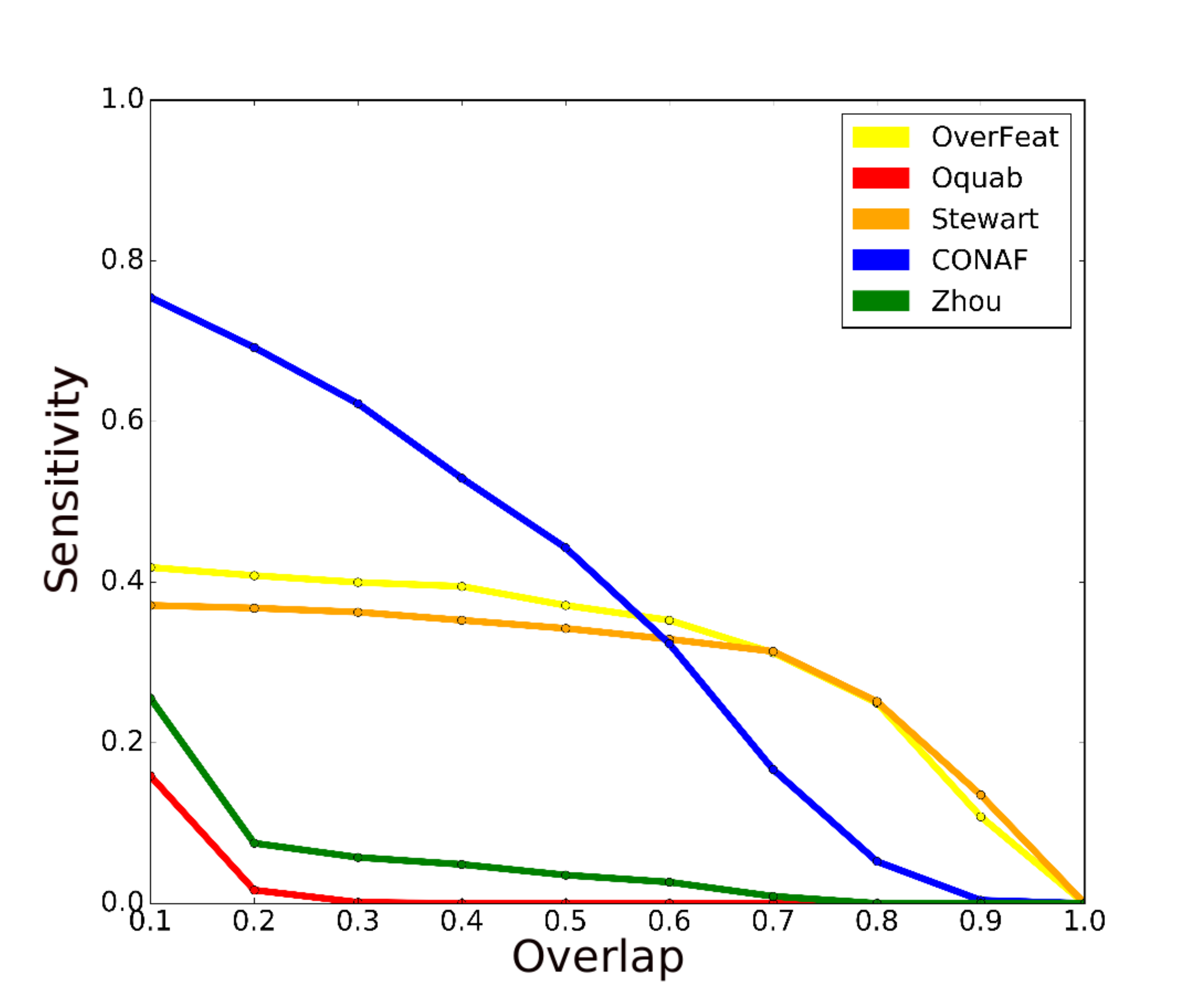}
			\captionsetup{labelformat=empty}
		\end{minipage}%
		\begin{minipage}[b]{.49\textwidth}
			\includegraphics[width=1\linewidth]{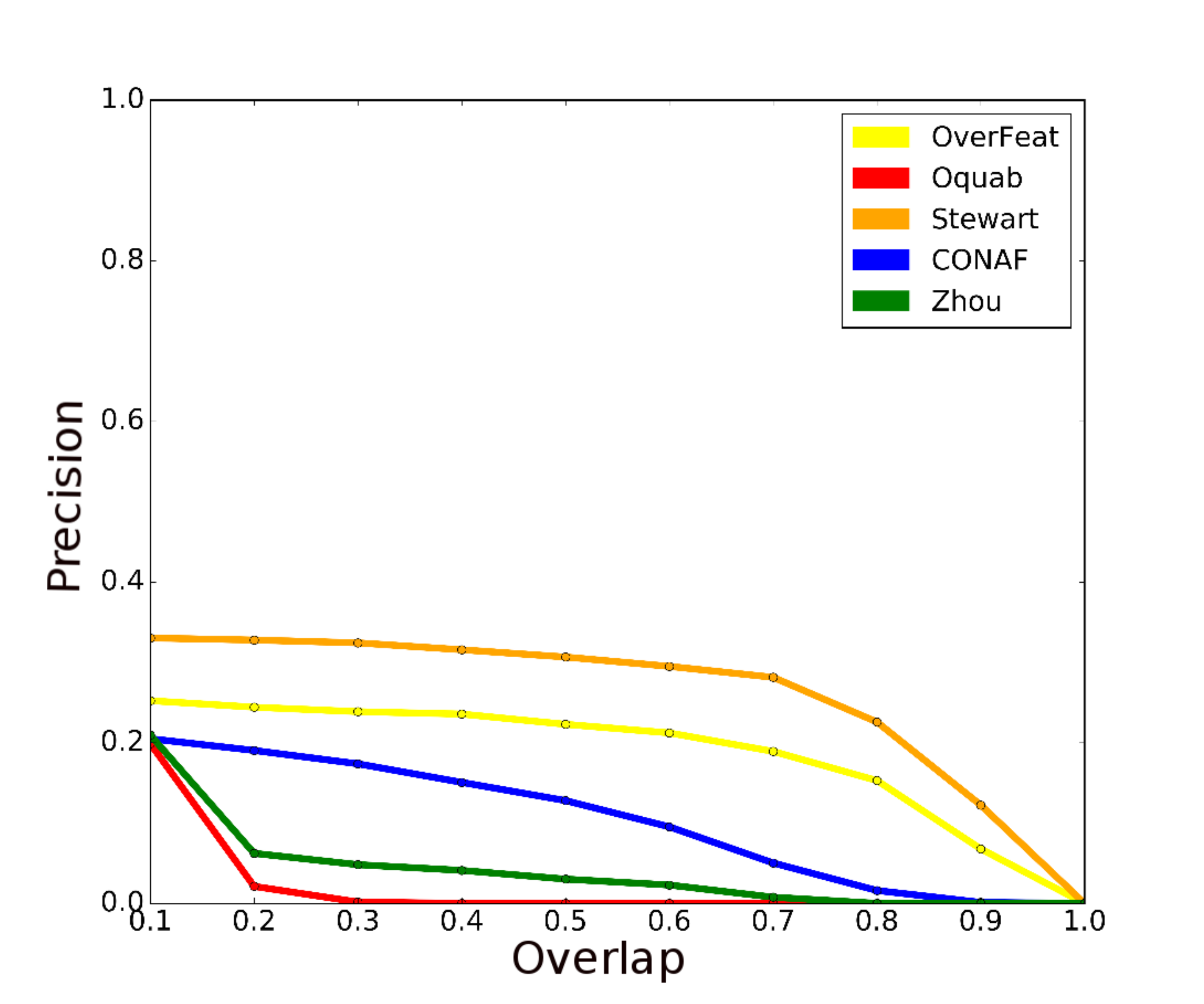}
			\captionsetup{labelformat=empty}
		\end{minipage}%
		\caption{CONAF localisation performance: recall (left) and precision (right) rates as function of the overlap threshold.}
		\label{fig:performance_overlap}
	\end{figure*}

	Table~\ref{table:locresult} summarizes all the localisation results. The table shows that, in terms of sensitivity and average overlap, CONAF achieves superior performance while OverFeat achieves the best precision. Furthermore, Fig.~\ref{fig:results_1} provides two examples comparing the localisation results obtained by CONAF and \citep{zhou2015learning}, which is best competitor study if looking at F1 score shown in Table \ref{table:normal_vs_all}. It can be noticed that the bounding boxes predicted by CONAF are closer to the ground truths in terms of location and shape, and in respect to the other methods.  Fig.~\ref{fig:performance_overlap} illustrates the relationship between the overlap threshold and sensitivity/precision for a number of competing algorithms. CONAF is capable of greater sensitivity than all other methods, whereas in terms of sensitivity/precision is in between the weakly-supervised and the object detection methods. 
	
	No comparable localisation metrics can be obtained using the RAM/RAMAF. Instead, we measure the percentage of regions contained within the bounding boxes that overlap with at least one of the \enquote{glimpses} taken by these models. In our experiments, RAMAF detected $82\%$ of the overall bounding boxes in the test set while the RAM model detected only $55\%$. This result indicates that RAMAF is capable to make a proper use of the additional spatial information that is accessible for a subset of the images. Additional and noticeable advantages have also been observed in terms of convergence rate; Fig. \ref{fig:ram_ramaf_av_acc} shows that RAMAF learns approximately five times faster compared to RAM.
	
	\section{Discussion and conclusions}
	
	Wherea as other imaging modalities for cancer detection (e.g. mammograms and the breast screening programme more widely) are routinely double-read and associated with an improvement in sensitivity of detection \citep{samRefA}, the same is not feasible with chest radiographs (due to the sheer volume of scans, $40\%$ of the $3.6$ billion annual medical images are chest radiographs) and a lack of resources. Machine learning systems powered by deep learning algorithms offer a mechanism to automate the second-reading process, but require large volumes of manually curated examples in order be trained, a process which is expensive and time-consuming. Furthermore, the automated detection of pulmonary lesions is a challenging task because nodules have a high variability in size and shape. Access to larger-scale radiological datasets has only recently enabled the joint modelling of images and radiological reports for automated screening purposes \citep{DBLP:journals/corr/ShinRLDYS16, Shin2016, Wang17}. This paper leverages a large database of NLP-labelled chest radiographs generated over a period of years in two large UK hospitals to explore the performance of different pattern classification algorithms to detect chest radiographs with lung lesions. A number of approaches, from weakly supervised learning to fully supervised object detection, have been compared with the purpose of improving the image classification task. To the best of our knowledge, this is the largest study to date exploring the potential of deep learning for pulmonary lesion detection. It is also the only study to use a heterogeneous historical database, comprising of all x-rays from over forty different scanners (including portable and stationary devices), and a well-diversified adult patient population. 
	
	Two novel methods have been proposed, undergirded by the principle that a significantly large proportion of weakly-labelled images can be combined with a smaller subset of manually annotated images through a visual attention mechanism in order to boost the classification performance. The idea of attention in deep neural networks is inspired by the human visual attention system. Spatial attention allows humans to selectively process visual information through prioritisation of an area within the visual field \citep{Rensink00} and significantly improve both recognition and detection performance, especially in images with cluttered background \citep{Cichy14}. Following the same principle, neural networks can be trained to focus on specific portions of an input signal that appear to be more strongly related to the task at hand. In CONAF, a localisation loss function is derived from inferred saliency maps and is combined with a traditional classification error to improve the overall performance. This architecture implements a supervised attention feedback mechanism since the error signal from the localisation component is used to further refine the saliency maps generated from the convolutional layers in a weakly supervised way. 
	
	CONAF can be interpreted as a type of feedback neural network \citep{Cao15, Stollenga14, Zamir17}, a recurrent architecture that iteratively uses high level features to back refine low level features and focus on the most salient image regions. Feedback neural networks without recurrent connections have been used recently for human pose estimation \citep{Carreira16}, where a self-correcting model progressively changes the initial prediction by iteratively feeding back the error predictions. In \citet{Newell16}, a stacked hourglass network is proposed to introduce bottom-up, top-down inference across multiple scales. In other domains, it has also been shown that network feedbacks can improve the task of locating human face landmarks \citep{Hu16}. Models implementing {\it soft attention mechanisms} typically learn by processing the entire input images using DCNNs. During learning, these models focus on certain parts of an input image that are directly associated with the demands of the task. The key idea is to learn features from a weighted average of all image locations where locations are weighted based on the saliency maps produced by the highest convolutional layers of the network. The intuition behind these approaches is that the saliency maps generated by the last convolutional layer of DCNNs trained on weakly labelled images highlight which regions of an image are important for classification. Soft attention has been used for learning a mapping between radiological reports and the corresponding histopathology specimens \citep{Zhang2017MDNET}. 
	
	The second model proposed here, RAMAF, uses a recurrent attention model with spatial feedback rewards to explore the image, building on previous work on chest radiographs \citep{Ypsilantis17}. While CONAF outperforms other state-of-the-art methods, RAMAF provides an improvement on the original approach when annotated images are available. RAMAF is an instance of {\it hard attention mechanisms} whereby learning evolves by iteratively focusing on selectively chosen regions within an image. In early attempts to introduce hard attention, the local information extracted from images was sequentially integrated in a variety of ways, e.g. through Boltzmann machines (BM) \citep{Denil11,Larochelle10} and geometric means of intermediate predictions \citep{Ranzato14}. More recent proposals have focused on stochastic exploration of a sequence of image regions. The number of computational operations involved in these models is independent of the size of the input image, in contrast to soft attention models whose computational complexity is directly proportional to the number of image pixels. While this allows hard attention models to scale up to large input images, the stochastic selection of image regions does not yield differentiable solutions, which hinders the applicability of back-propagation. Instead, these models are typically trained using reinforcement learning methods \citep{Mnih14, Williams92}. 
	
	In comparison to other methods, using the F1 score – calculated from precision and sensitivity – our image classification results are an improvement over other documented methodologies. By combining the large set of reported images with a high quality subset of annotated lesions, we show that the sensitivity can be improved whilst attaining an acceptably low level of false postives, which is essential for clinical use. When investigating lesion localisation, CONAF achieves a much higher sensitivity compared to other algorithms. OverFeat and Stewart's method, which are trained using object detection, can achieve higher precision, but at the cost of a much lower sensitivity. Moreover, CONAF achieves very good localisation performance, i.e. a very high degree of overlap between the predicted lesion and the manually identified ground truth regions.
	
	In the literature, existing CAD systems for pulmonary lesion detection have been tested on datasets with sample sizes up to hundreds of patients \citep{bushlung,moore2011sensitivity, szucs2013comparison}. More recently, access to large number of historical exams has allowed studies to be scaled up to several thousand examples \citep{openi, Wang17}. For chest x-rays, a database of $7,284$ images spanning thirteen disease classes (including $211$ lesion examples and $1,379$ normal examples) has recently been used to automatically learn to detect a disease and annotate its context \citep{DBLP:journals/corr/ShinRLDYS16}. More recently, a database of $108,948$ chest radiographs, spanning eight disease classes, has been made publicly available (with $1,971$ examples of lesions and $84,312$ normal examples) \citep{Wang17}. Direct comparisons with published results are potentially misleading because of noticeable differences in how comparisons have been done (e.g. whether normal exams are compared to exams with lung lesions only, rather than including the full spectrum of abnormalities that are typically observed). Our empirical results are particularly promising considering that the image labels used in this study are, inevitably, noisy. Several recent studies in other domains have shown that deep convolutional neural networks for image classification are sufficiently robust against noisy labels ~\citep{GuanGDH17,Rolnick17}. 
	
	In future work, the simple network architectures describe here could be further improved. In particular, instead of using a fixed number of glimpses, RAMAF could be extended to adaptively decide how much context is required in order to correctly classify each image (e.g. the size and the number of glimpses). Such as extension could reduce the computational time and add an additional layer of interpretability.
	
	\section*{Acknowledgments}
	Giovanni Montana acknowledges financial support of a King's Health Partner's R\&D Challenge Fund award as well as a King's Health Accelerator Fund award.
	
	\section*{References}
	
	\bibliographystyle{abbrvnat}
	\bibliography{refs}

	\newpage
	\appendix
	
	\renewcommand\thefigure{\thesection.\arabic{figure}}  
	\renewcommand{\thetable}{A\arabic{table}}%
	
	\section{The NLP system for automated image tagging} \label{appendix:nlp}
	\setcounter{figure}{0} 
	\setcounter{table}{0} 
	
	All the radiological reports were analysed using a natural language processing (NLP) system that implements a combination of machine learning and rule-based algorithms for clinical entity recognition, negation detection and entity classification. This analysis identified $406,935$ exams with no reported evidence of lung lesion ($101,766$ of them identified as {\it Normal}), and $25,081$ exams containing a reference to lung lesions (class {\it Lesion}). Although the labels we used may occasionally be noisy due to reporting/human errors and/or NLP-related errors, our working assumption when training the proposed computer vision architectures was that the majority of the labels were accurate. The NLP system we developed and used in this study is composed by four stages which are described below.  
	
	\subsection{Entity detection}
	
	At a first stage, the NLP system process each radiological report and automatically identifies medical concepts, or {\it entities}, using sources of information: RadLex \citep{radlex}, a radiology ontology, and MeSH \citep{mesh}, a general medical ontology. RadLex and MeSH are hierarchically structured lexicons for radiological and general medical terminology, respectively. Additionally, at this stage, the hierarchical structure of these lexicons is used to associate each identified entity to one of four semantic classes: \emph{Clinical Finding, Body Location, Descriptor} and \emph{Medical Device}. \emph{Clinical Finding} encompasses any clinical-relevant radiological abnormality, \emph{Body Location} refers to the anatomical area where the finding is present, and the \emph{Descriptor} includes all adjectives used to describe all the other classes. The \emph{Medical Device} class is used to label any medical apparatus seen on chest radiographs, such as pacemakers, intravascular lines, and nasogastric tubes.    
	
	Initially, each sentence in a report is tokenised, split using the Stanford CoreNLP suite \citep{Manning14}, converted to lower case and lemmatised using NLTK \citep{Bird09}. An attempt is then made to match the longest possible sequence of words, a target phrase, to a concept name in RadLex \citep{Langlotz06} and Mesh \citep{NLM16}. For example, the entity \enquote{enlarged heart} can be associated with the controlled vocabulary concept \enquote{cardiomegaly}. When a match is successful, the target phrase is annotated with the corresponding concept thus creating an {\it entity}. When no match is found, the algorithm attempts to look up the target phrase in the English Wikipedia redirects database. When a match is found, the name of the target Wikipedia article is checked against the name of RadLex/MeSH concepts (e.g. {\it oedema} redirects to {\it edema} in RadLex). All string matching operations are performed using \emph{SimString} \citep{Okazaki10} using a \emph{cosine} similarity measure with a similarity threshold value of $0.85$. This allows to match misspelt words, e.g. {\it cardiomegally} to the correct concept {\it cardiomegaly}.
	
	\subsection{Negation detection}

	At the second stage, a negation attribute is assigned to each entity indicating whether the entity is negated or affirmed. For this stage, the NLP system first obtains the NegEx predictions \citep{Chapman13} for each of the entities identified in the first step. Next, the system generates a graph of grammatical relations as defined by the Universal Dependencies \citep{stanfordparser} from the Stanford Dependency Parser. It then removes all the relations in the graph except the negation relation and the \emph{or} disjunction. Given the NegEx and the reduced dependency graph, the system finally classifies an entity as negated if any of the following two conditions are found to be true: (1) any of the words that are part of the entity are classified as negated or in a \emph{or} disjunction relation with another word that is in a negation relation; (2) if an entity is classified by NegEx as negated, it is the closest entity to negation trigger and there is no negation relationship in the sentence. If none of the above conditions are true, then the entity is classified as affirmed. This approach is similar to DEEPEN \citep{Mehrabi15} with the difference that the latter considers all first-order dependency relations between the negation trigger and the target entity.

	\subsection{Relation classification}

	In the third step, the NLP system identifies the {\it semantic relations} between pairs of entities, which are eventually used to identify radiological classes in the reports. The system considers two types of directed relations: \enquote{\emph{located in}} and \enquote{\emph{described by}}. We impose the restriction that a relation can only exist between entities found in the same sentence. In addition, the relationship between entities are limited according to the semantic class assigned to each entity. Therefore the relation \enquote{\emph{located in}} between two entities, denoted as $e_{1}, e_{2}$, can only exist if $e_1$ is a \emph{Clinical Finding} or \emph{Medical Device} and $e_2$ is a \emph{Body Location}. The relation  \enquote{\emph{described by}} can only exist if $e_1$ is a \emph{Clinical Finding}, \emph{Medical Device} or \emph{Body Location} and $e_2$ is a \emph{Descriptor}.

	To identify each relation type, we train a separate binary classifier based on a CNN model \citep{nguyen2015relation}. At prediction time the model receives as input a sentence and classifies a single candidate relation as true or false. Each input sentence is represented by a vector of embeddings that corresponds to the tokens in the sentence, preserving the order. In addition, the model receives as input position features that encode the relative distance of each token in the sentence to the arguments of the candidate relation. The CNN architecture is as follows. The word embeddings and the position features are concatenated and passed as input to two convolutional layers, where each layer is followed by a max pooling layer. Then, the output of the convolutional and max pooling layers is passed as input to two fully connected layers where each one is followed by a dropout layer. Finally, a softmax layer is applied for binary classification.

\begin{figure}[H]
		\centering
		\subfigure{
			\includegraphics[width=0.9\linewidth]{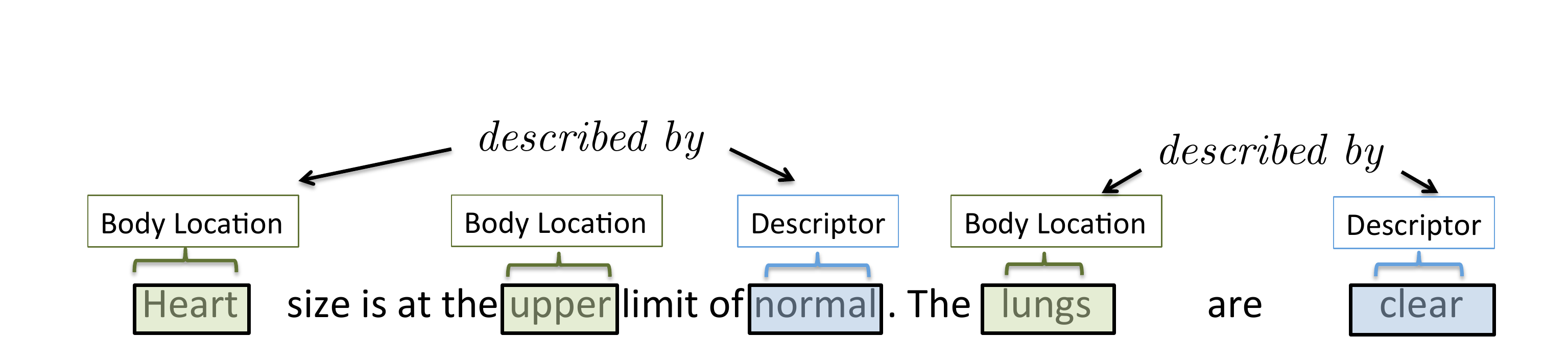}
		}\\
		(a)
		\subfigure{
			\includegraphics[width=0.9\linewidth]{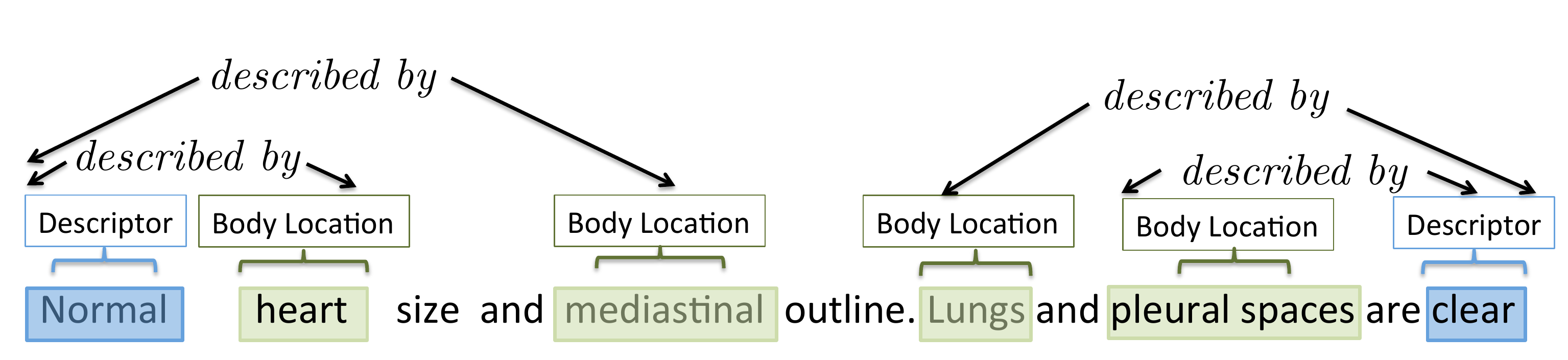}
		}\\
		(b)
		\subfigure{
			\includegraphics[width=0.9\linewidth]{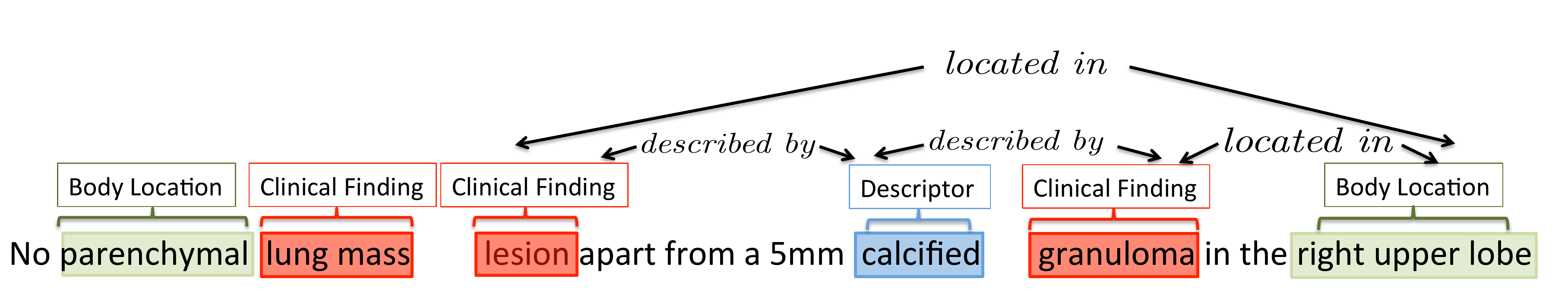}
		}\\
		(c)
		
		\subfigure{
			\includegraphics[width=0.9\linewidth]{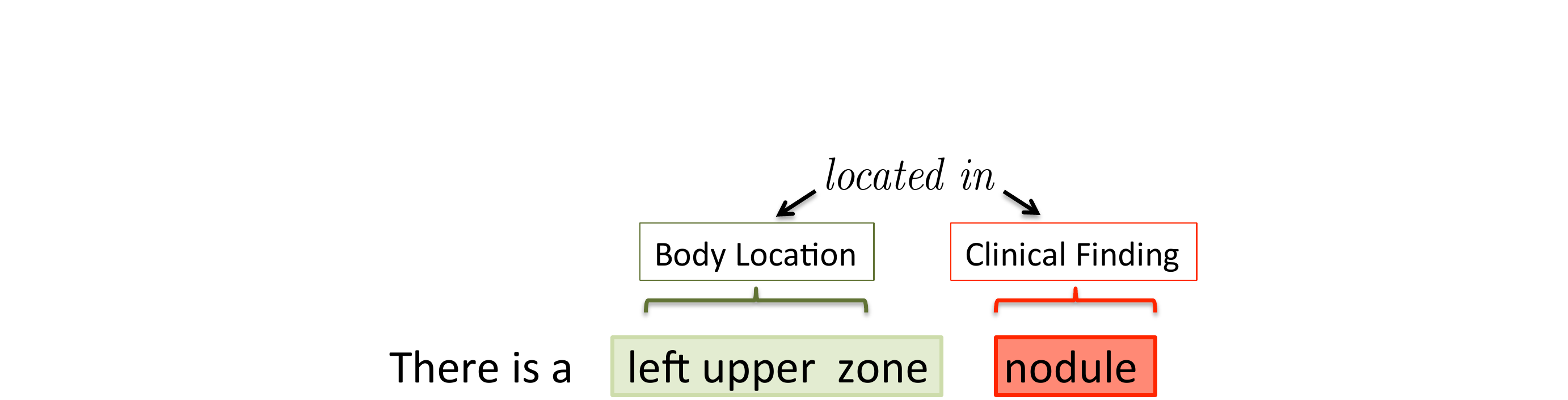}
		}
		(d)
		\subfigure{
			\includegraphics[width=0.9\linewidth]{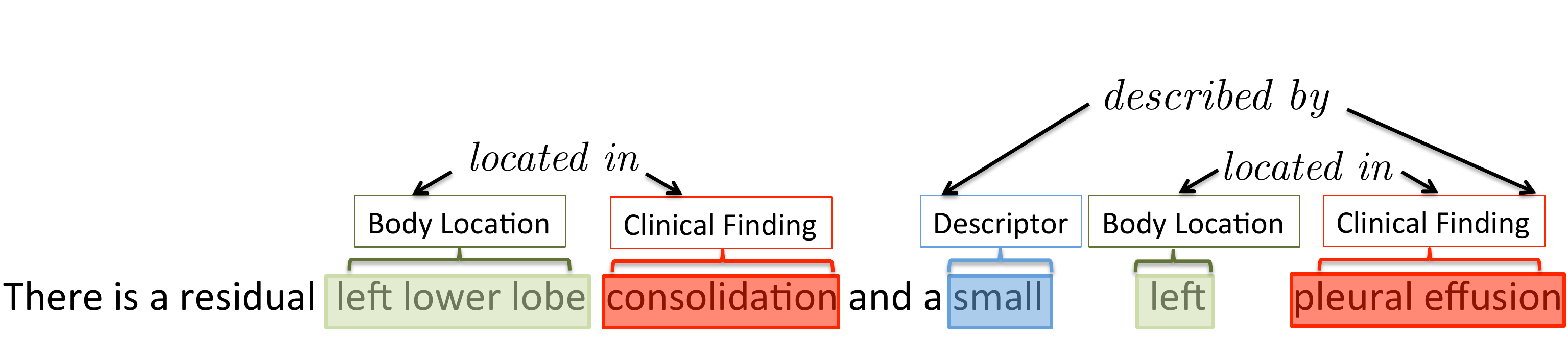}
		}
		(e)
		\caption{Five examples of a radiological report annotated by the NLP system. (a), (b) reports were classified as {\it Normal}, (c), (d) reports were classified as {\it Lesion} and (e) was classified as pleural effusion and consolidation. The pleural effusion and consolidation were included in the class \emph{Others} under the diseases of Pleural Effusion/Abnormality and Airspace Opacifation respectively (see TABLE~\ref{tab:NLP_results}).}
		\label{fig:nlp_example}
	\end{figure}

	The dataset used for the \enquote{\emph{located in}} relation type consisted of $1,100$ relationships of which $729$ were annotated as {\it true} and $371$ were annotated as {\it false}. The corresponding dataset for the \enquote{\emph{described by}} classification model had $507$ {\it true} and $593$ {\it false} relations. The maximum distance between the relation arguments were limited to $16$ words which was also the maximum limit of the input sentence length. All candidate relations with arguments more than $16$ words apart were automatically classified as false. As loss function we used the cross-entropy between the predicted probabilities of existence/absence of the relation and the true labels from the manual annotation. The CNN was trained on a GPU for $50$ epochs in batches of $5$ sentences using SGD with momentum and with learning rate set to $0.005$. The word embeddings used as input during training and prediction time were obtained by training the GloVe model \citep{Pennington14} on $743,480$ radiology reports. The embedding size was set to $20$. Using a larger embedding size for a relative small vocabulary used by radiologists provided no performance benefits. An example of an automatically annotated radiological report is illustrated in Fig. \ref{fig:nlp_example}. It can be seen that the NLP automatically associates each identified entity to one of the four semantic classes and identifies the semantic relations between the pairs of entities. 
	
	\subsection{Classification}
	
	In the final stage, the NLP system labels the reports by using a ruled-based approach for classification. It processes one report at a time taking as input the list of all entities, negation attributes and relations extracted in previous stages. The system checks the entities and relations from the input report against a list of rules. When a rule is activated then the report is labelled with the radiological class corresponding to the matching rule. If the report does not match any rule, it is not be labelled and remaining unclassified. Overall, the system uses $826$ rules, each one mapping to one of the radiological classes, which were carefully designed in close collaboration with expert radiologists. 
	
	\subsection{Validation study}\label{Supp:validationstudy}
	
	To verify the performance of the NLP system, a subset of $4,652$ randomly selected reports was independently labelled by two radiologists, blinded to the images. Approximately $ 7\% $ of these exams were labelled as \emph{Lesion}, $15\% $ as \emph{Normal} and the rest as \emph{Others}. Table~\ref{table:NLPperformance} shows the performance of the NLP system on these exams. It should noted that good performance has been achieved overall, in particular for \emph{Normal} exams.

	\begin{table}[H]
		\centering 
		\caption{NLP performance measures by average accuracy, F1 measure, sensitivity, specificity, precision and negative predictive value (NPV).}
		\scalebox{0.9}{
			\begin{tabular}{|c|c|c|c|c|c|c|}
				\hline   Class &  F1 & Sensitivity & Specificity & Precision & NPV \\ 
				\hline  {\it Lesion}  & 0.77 & 0.77 & 0.98 & 0.76 & 0.98 \\    
				\hline  {\it Normal}  & 0.98 & 0.97 & 0.99 & 0.99 & 0.99 \\
				\hline  {\it Others}  & 0.99 & 0.99 & 0.58 & 0.77 & 0.38 \\
				\hline 
			\end{tabular}
		}		 
		\label{table:NLPperformance}
	\end{table}
	
	In Table~\ref{tab:NLP_results} we summarize the NLP performance results by F1 score, sensitivity, specificity, precision and (NPV) for all the available diseases which form the class \emph{Others}. The percentage of scans that contain a specific disease is  given in the column called Prevalence. It can be noticed that NLP system achieves very good performance across all available diseases.

	\begin{table}[H]
		\begin{center}
			\caption{NLP performance results by precision, sensitivity, Specificity, F1 score and negative predictive value (NPV) across all the available diseases that form the class \emph{Others}. The Prevalence represents the percentage of manually validated exams that contain a specific disease.}
			\label{tab:NLP_results}
			\scalebox{0.75}{
				\begin{tabular}{|l|c|c|c|c|c|c|c|}
					\hline \swapcol{3}{6}  Class						& Prevalence	& Precision	& Sensitivity	& Specificity	& F1		& NPV	\\
					\hline  \swapcol{3}{6} Abnormal Other				& 15.37 \%				& 0.9690	& 0.9748		& 0.9820		& 0.9719	& 0.9854	\\
					\hline  \swapcol{3}{6} Airspace Opacifation		& 17.90 \% 			& 0.9418	& 0.8965		& 0.9795		& 0.9186	& 0.9625	\\
					\hline  \swapcol{3}{6} Bone Abnormality			& 1.92 \%			& 0.9000	& 0.9329		& 0.9919		& 0.9162	& 0.9948	\\
					\hline  \swapcol{3}{6} Cardiomegaly				& 11.44 \%			& 0.9959	& 0.9939		& 0.9995		& 0.9949	& 0.9993	\\
					\hline  \swapcol{3}{6} Collapse					& 1.86 \%			& 0.9448	& 0.9716		& 0.9931		& 0.9580	& 0.9965	\\
					\hline  \swapcol{3}{6} Hiatus Hernia				& 0.86 \%		& 0.9771	& 0.9922		& 0.9993		& 0.9846	& 0.9998	\\
					\hline  \swapcol{3}{6} Interstitial Shadowing		& 2.99 \%		& 0.9964	& 0.8671		& 0.9998		& 0.9272	& 0.9902	\\
					\hline  \swapcol{3}{6} Intra-abdominal Pathology	& 0.33 \%			& 0.9289	& 0.8756		& 0.9968		& 0.9015	& 0.9940	\\
					\hline  \swapcol{3}{6} Medical device				& 32.87 \%			& 0.9852	& 0.9434		& 0.9927		& 0.9639	& 0.9713	\\
					\hline   \swapcol{3}{6}Paratracheal Hilar Enlargement	& 0.72 \%		& 0.8519	& 0.8880		& 0.9907		& 0.8696	& 0.9932	\\
					\hline   \swapcol{3}{6} Pleural Effusion/Abnormality	& 20.97 \%		& 0.9790	& 0.9039		& 0.9943		& 0.9399	& 0.9725	\\
					\hline   \swapcol{3}{6} Pneumomediastinum			& 0.09 \%	& 0.8700	& 0.9560		& 0.9971		& 0.9110	& 0.9991	\\
					\hline  \swapcol{3}{6} Pneumothorax				& 5.76 \%		& 0.7707	& 0.9688		& 0.9805		& 0.8585	& 0.9979	\\
					\hline  \swapcol{3}{6} Subcutaneous Emphysema		& 0.34 \%		& 0.9677	& 0.9615		& 0.9989		& 0.9646	& 0.9986	\\			
					\hline 
				\end{tabular}
			} 
		\end{center}
	\end{table}

	\section{The RAMAF model} \label{appendix:ramaf}
	\setcounter{figure}{0} 
	\setcounter{table}{0} 
	
	The model is trained to infer a stochastic policy which is optimal with respect to the rewards or returns the model can expect when interacting with the radiographs. This can be seen as a reinforcement learning task in a partially observable Markov decision problem (POMDP). We task consists of learning a stochastic policy representation $\pi(\textbf{s}_{t}| \textbf{S}_{1:t}; \pmb \theta)$ with an internal memory which maps the sequence of \enquote{glimpses} $\textbf{S}_{1:t}$ to a distribution over actions for the current step $t$. We define the policy $\pi$ as RNN with long short-term memory (LSTM) units \citep{Hochreiter97} where the information from previous glimpses $\textbf{S}_{1:t}$ is summarized in the hidden state $\textbf{h}_{t}$. The policy of the model $\pi$ induces a distribution over possible interaction sequences $\textbf{S}_{1:T}$ and we aim to maximize the reward under this distribution:
	\begin{equation}
	J(\pmb \theta) = \mathbb{E}_{p(\textbf{S}_{1:T}; \pmb \theta)}[R(\textbf{S}_{1:T})],
	\end{equation}
	where $p(\textbf{S}_{1:T};\pmb \theta)$ represents the probability of the sequence $\textbf{S}_{1:T}$ and depends on the policy $\pi$.
	
	Computing the expectation exactly is non-trivial since it introduces unknown environment dynamics. Formulating the problem as a POMDP allows us to approximate the gradient using an algorithm known as REINFORCE \citep{Williams92}:   
	
	\begin{equation}
	\nabla_{\pmb \theta} J \approx \frac{1}{N}\sum_{i=1}^{N}\sum_{t=1}^{T}\nabla_{\pmb \theta} \log \pi (\textbf{s}_{i,t}|\textbf{h}_{i,t-1})R(\textbf{S}_{i,1:T}).
	\end{equation}
	Eq. (3) requires us to compute $\nabla_{\pmb \theta} \log \pi (\textbf{s}_{i,t}|\textbf{h}_{i,t-1})$, but this is the gradient of the RNN that defines our model evaluated at time step $t$ and can be computed by backpropagation \citep{Wierstra07}. A well-known problem with the Monte Carlo approach is the often high variance in the estimation of the gradient direction resulting in slow convergence \citep{Marbach03,Peters06}. One way to solve this problem and reduce the variance is to include a constant baseline reward $b$ (first introduced by Williams \citep{Williams92}) into the gradient estimate: 
	\begin{equation}
	\nabla_{\pmb \theta} J \approx \frac{1}{N}\sum_{i=1}^{N}\sum_{t=1}^{T}\nabla_{\pmb \theta} \log \pi (\textbf{s} _{i,t}|\textbf{h}_{i,t-1})[R(\textbf{S}_{i,1:T}) - b_{i}].    
	\end{equation}
	We select $b_{i}= \mathbb{E}_{\pi}[R(\textbf{S}_{i,1:T})]$ \citep{Sutton00} and learn it by reducing the squared error between $R(\textbf{S}_{i,1:T})$ and $b_{i}$ \citep{Mnih14}. The resulting algorithm increases the log-probability of an action that was followed by a larger than expected cumulative reward, and decreases the probability if the obtained cumulative reward was smaller.
	
	We use the above algorithm to train the model when the majority of the best actions (e.g. locations) within the X-ray image are unknown and only a very small number of parenchymal lesion locations are provided. In our problem we know the labels of the X-ray images and therefore we can optimize the policy to output the correct label at the end of the observation sequence $\textbf{S}_{1:T}$. This can be achieved by maximizing the conditional probability of the true label given the observations from the image. Consistent with \citep{Mnih14}, we optimize the cross entropy loss to train the network to correctly classify the X-ray images. Also we train the part of the model which propose the observation locations (locator) using the algorithm described above.
	
\begin{figure}[H]
		\centering
		\includegraphics[scale=0.45]{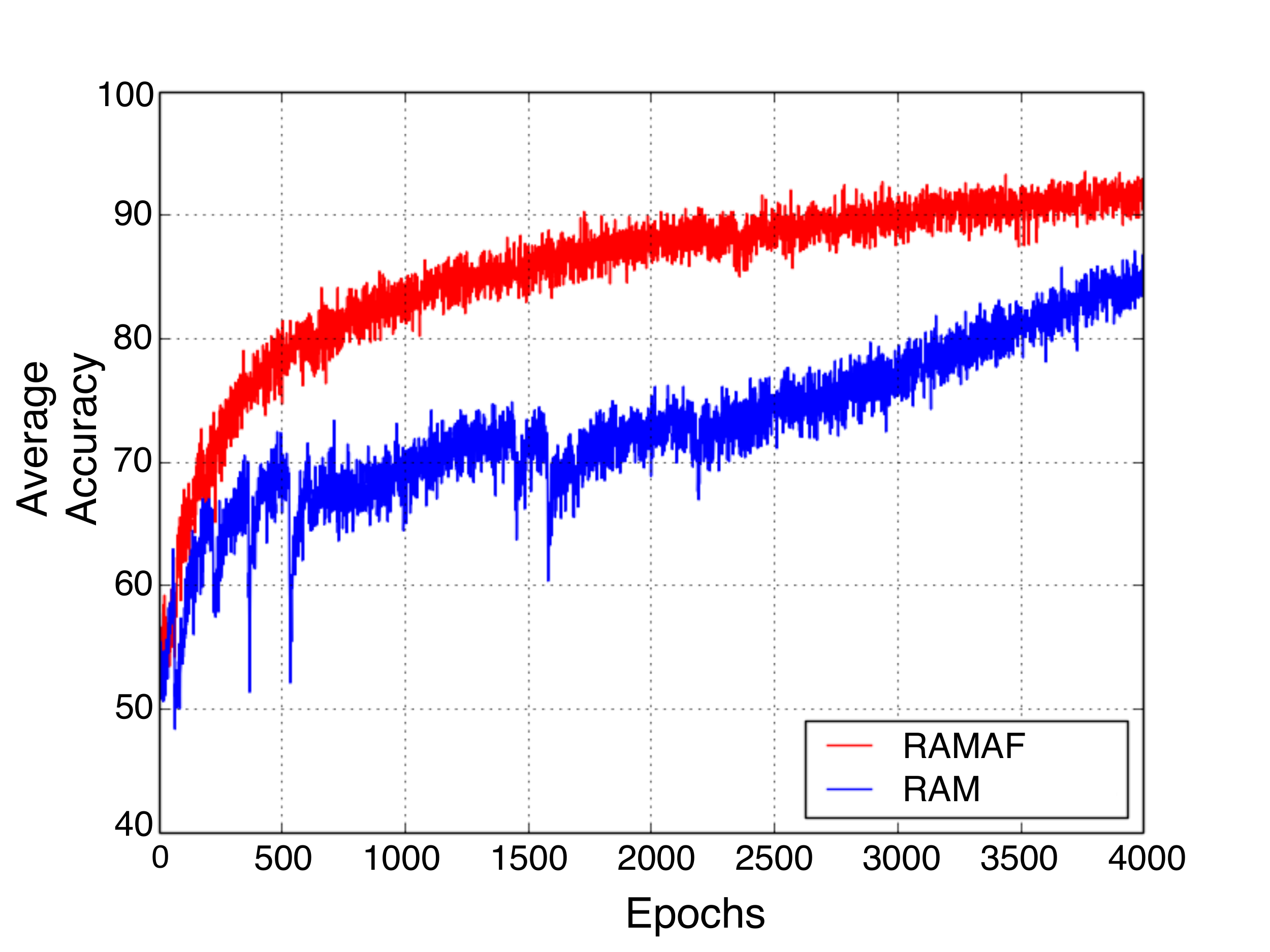}
		\caption{Training average accuracy for the RAM and RAMAF models. RAMAF learns $5$ times faster compared to RAM.}
		\label{fig:ram_ramaf_av_acc}
	\end{figure}

	Fig. \ref{fig:ram_ramaf_av_acc} illustrates that RAMAF learns approximately five times faster compared to RAM. The spatial reward provided by the limited number of annotated bounding boxes forces to model to attend the regions that are likely to contain a lesion at a faster rate. In contrast, RAM does not use any spatial reward, and thus ends up spending more time exploring irrelevant image portions initially.

	\begin{figure}[H]
		\centering
		\includegraphics[width=1\linewidth]{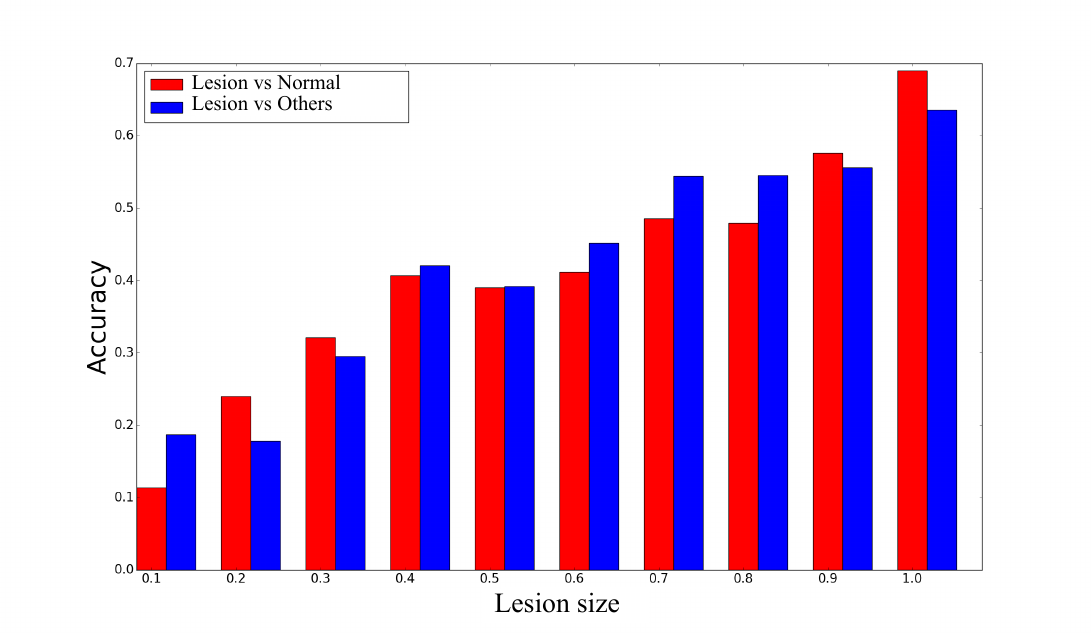}
		\caption{Accuracy of CONAF model by lesion size for both experiments. Lesions have been grouped by size into deciles, with 1.0 representing the top decile (largest masses); and 0.1 representing the first decile (smallest nodules).}
		\label{fig:results_3}
	\end{figure}
	
\end{document}